\documentclass[runningheads]{llncs}
\usepackage[T1]{fontenc}
\usepackage{graphicx}
\usepackage{xcolor}
\usepackage{subcaption}

\begin{document}
\title{How a Model’s Architecture and Performance Influences Explainability: A Study of Grad-CAM and CRAFT in CNN-Based Image Classification}
\titlerunning{How a Model’s Architecture and Performance Influences Explainability}

\author{Kilian Göller$^*$\orcidID{0009-0006-3522-2246} \and
Haadia Amjad$^*$\orcidID{0009-0001-9227-9496} \and
Steffen Seitz\orcidID{0000-0002-8389-8869} \and
Carsten Knoll\orcidID{0000-0001-7462-771X} \and
Ronald Tetzlaff\orcidID{0000-0001-7436-0103}
}

\authorrunning{K. Göller et al.}
%
\institute{
Fundamentals of Electronics, TUD Dresden University of Technology, Dresden, Germany\\
\email{\{kilian.goeller, haadia.amjad, steffen.seitz, carsten.knoll, ronald.tetzlaff\}@tu-dresden.de}}
\maketitle              
{\footnotesize $^*$Equal contribution.}


\begin{abstract}
Deep learning models, particularly convolutional neural networks (CNNs), are widely adopted in computer vision, yet their opaque decision-making raises concerns in high-risk applications like healthcare and autonomous systems, where trust and accountability are critical. Therefore, robust explainable AI (XAI) methods are desirable to provide transparent insights into model decisions. To ensure these methods are reliable and effective across diverse scenarios, it is important to understand how model performance and architectural complexity influence explanation quality. Our study takes an initial step toward systematically investigating the impact of these two factors on key XAI evaluation metrics. We constructed a pipeline using 10 subsets of ImageNet-1K, each containing 80 unique classes, on each of which we trained three CNN architectures of increasing structural complexity: VGG16, ResNet50, and ConvNeXt-Tiny. For each individual model, we analyzed 10 training checkpoints representing increasing levels of model accuracy using two XAI methods: Grad-CAM (feature-attribution-based) and CRAFT (concept-based). We assessed these methods across four evaluation categories: \textit{Faithfulness}, \textit{Correctness}, \textit{Sensitivity}, and \textit{Simplicity}. Our findings reveal that as training progresses, the complex architectures (ResNet50 and ConvNeXt-Tiny) exhibit improved \textit{Faithfulness} and lower \textit{Sensitivity}, indicating more \textit{robust} explanations. In contrast, VGG16’s simpler architecture yields \textit{simpler} but less \textit{faithful} and \textit{correct} explanations.

\end{abstract}

\section{Introduction}
Despite recent advances in transformer-based models, convolutional neural networks (CNNs) remain widely used in computer vision due to their parameter efficiency on smaller datasets and architectural simplicity. Classic architectures such as VGG \cite{simonyan2015deep} and ResNet \cite{he2016deep} continue to be widely utilized \cite{liu2024cancer,rath2025resnet}, while modern designs like ConvNeXt \cite{liu2022convnet} demonstrate competitive performance against state-of-the-art transformer models (especially on small-scale datasets), underscoring their ongoing relevance.

\noindent However, the black-box nature of CNNs poses serious challenges in high-risk domains such as healthcare \cite{agarwal2023automated}, autonomous systems \cite{zhao2025autonomous}, and fault detection in industry \cite{seitz2024towards}, where transparency, trust, and accountability are essential. Explainable Artificial Intelligence (XAI) addresses these issues by providing interpretable insights into model decisions \cite{kalasampath2025literature}. Evaluating the quality and reliability of these explanations is therefore critical, as it depends not only on the XAI method but also on the underlying model architecture and its performance.

\noindent XAI methods are commonly divided into feature-attribution approaches (e.g., assigning importance to input pixels) and concept-based methods (using higher-level, human-interpretable concepts) \cite{wang2024gradient,poeta2023concept}. To assess explanation quality, we adopt four evaluation categories from Nauta et al. \cite{nauta2023anecdotal} and Lopes et al. \cite{lopes2022liteval}: \textit{Faithfulness} (how accurately do the explanations reflect the model’s decision-making?), \textit{Correctness} (how well do the explanations align with a model’s learned parameters?), \textit{Sensitivity} (how robust are the explanations to input perturbations?) and \textit{Simplicity} (how concise and focused is an explanation?).

\noindent Although recent work suggests that XAI performance may strongly depend on the underlying CNN architecture \cite{knoll2025xaiev}, the specific influence of architectural complexity and model accuracy on explanation quality remains underexplored.

\noindent In this study, we define \textit{architectural complexity} as the degree of structural sophistication (e.g., residual connections, advanced normalization, or transformer-inspired elements) beyond a basic sequential CNN. Thus, VGG16 represents a simple architecture, ResNet50 adds residual connections, and ConvNeXt-t incorporates further modern design choices.

\begin{figure*}[!htb]
    \centering
    \includegraphics[width=1\textwidth]{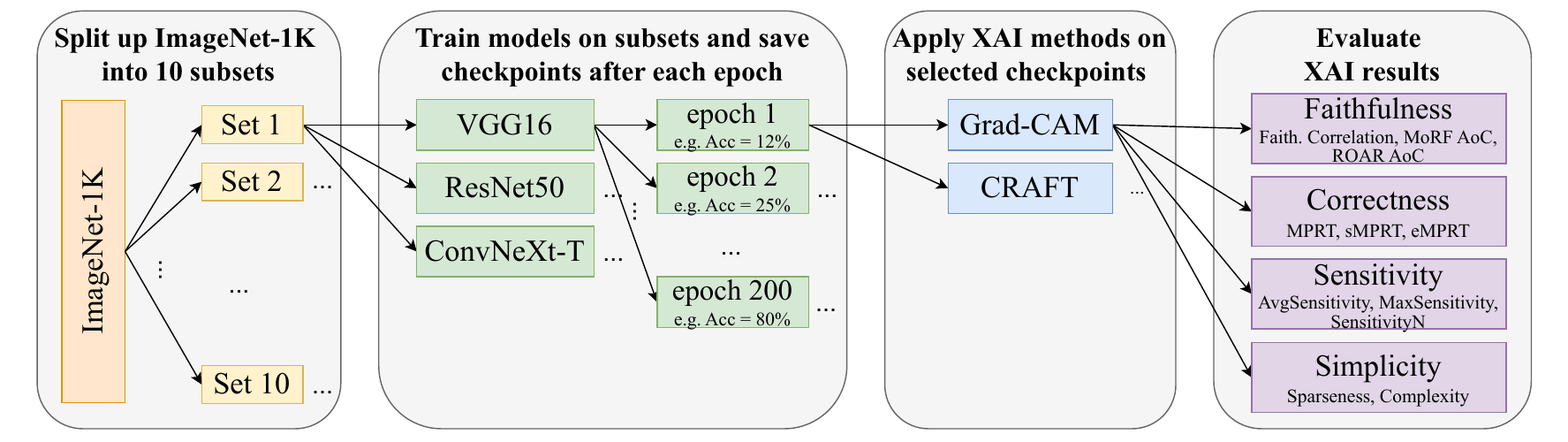}
    \caption{Applied methodology of this study: We split ImageNet-1K into 10 subsets, each with 80 unique classes. Three CNN architectures (VGG16, ResNet50, and ConvNeXt-Tiny) are trained on each subset. For each architecture, we select 10 checkpoints with varying accuracies and apply Grad-CAM and CRAFT, evaluating the explanations using \textit{Faithfulness}, \textit{Correctness}, \textit{Sensitivity}, and \textit{Simplicity} metric categories.}
    \label{fig:flowchart}
\end{figure*}

\noindent We investigate this interplay using the three forementioned CNNs of increasing complexity trained on 10 subsets of ImageNet-1K (each with 80 unique classes). For each architecture and subset, we select 10 checkpoints spanning a range of accuracies, apply the feature-attribution method Grad-CAM \cite{selvaraju2017gradcam} and the concept-based method CRAFT \cite{fel2023craft}, and evaluate the explanations across the four metric categories (see Fig. \ref{fig:flowchart}).

\noindent Our study yields the following novel empirical findings:
\begin{enumerate}
    \item We demonstrate that as training progresses, complex architectures (ResNet50 and ConvNeXt-t) improve in \textit{Faithfulness}, \textit{Correctness} and \textit{Sensitivity} but exhibit less \textit{Simplicity}, resulting in less focused and concise explanations.
    \item We show that, compared to complex architectures, simpler VGG16 models yield \textit{simpler} explanations (particularly at later training stages), but exhibit higher (usually undesired) \textit{Sensitivity} and less \textit{Faithfulness} and \textit{Correctness}.
\end{enumerate}

\noindent While our contributions do not claim to provide a comprehensive general proof applicable to all scenarios (in terms of CNN models, XAI methods, and datasets), they still represent a valuable step toward understanding the relations between model architecture, accuracy, and explanation quality.
\vspace{-\baselineskip}
\section{Related Work}
\noindent While many studies already highlighted the importance of balancing accuracy and explainability in different scenarios \cite{herm2023stop,petkovic2023trustworthy}, they lack systematic analysis of how model performance and model architecture actually impact explainability. To address this problem, recent work has begun to evaluate XAI methods more rigorously.

\noindent Klein et al. \cite{klein2024navigating} respond to inconsistencies in explainability evaluation by proposing a structured comparison of popular methods like SHAP \cite{lundberg2017unified}, LIME \cite{ribeiro2016should}, and Grad-CAM. Using model-based metrics (e.g., faithfulness, stability) and human-centered criteria (e.g., plausibility), they test methods across multiple datasets and architectures. Their findings show no clear winner, stating that XAI method performance varies by context. They emphasize the need to select XAI tools based on specific use cases and advocate for clearer evaluation standards, a conclusion echoed by Herm et al. \cite{herm2023stop}. However, they did not specifically investigate the influence of training dynamics or architectural complexity on explanation quality.

\noindent Terzieva et al. \cite{terzieva2023identifying} examine how feature attributions evolve during model training, using Grad-CAM and LRP \cite{LRP} across image classifiers. They track attribution changes across epochs, finding, through entropy metrics, that explanations are highly unstable early in training and only stabilize later, despite earlier stabilization of accuracy. This suggests that explanations during training may be misleading. However, their study focuses solely on feature-attribution-based methods and uses uncertainty instead of XAI quality metrics for evaluation. Also the impact of architectural differences between models was not part of their scope.

\noindent While previous work has studied the trade-off between accuracy and explainability, as well as the evolution of explanations during training, none of them has systematically investigated the combined effects of model performance, architectural complexity, and XAI method type on standardized evaluation metrics. Our study addresses this gap by analyzing the interplay of training progress (in terms of model accuracy), architectural complexity (sequential $\rightarrow$ residual $\rightarrow$ transformer-inspired, represented by VGG16, ResNet50, ConvNeXt-t), and XAI method type (feature-attribution-based Grad-CAM vs. concept-based CRAFT) across four evaluation categories using a comprehensive pipeline on ImageNet-1K subsets.

\section{Materials and Methods} \label{methods}
In this section, we provide details regarding the datasets, models, XAI methods and metrics used in our experiment.

\subsection{Datasets}
We use the ImageNet-1K dataset, a large-scale benchmark for image classification containing approximately 1.28 million training images and 50,000 validation images across 1,000 classes. To enable robust comparisons across multiple datasets while maintaining a consistent number of classes, we randomly divided ImageNet-1K into 10 disjoint subsets, each containing 80 unique classes. Each subset includes a training split with 1,300 images per class (104,000 images total per subset) and a test split with 50 images per class (4,000 images total per subset). This setup ensures multi-class classification tasks with sufficient data diversity for training and evaluation.

\subsection{Models}
\begin{figure}[htbp]
  \centering
  \includegraphics[width=0.7\textwidth]{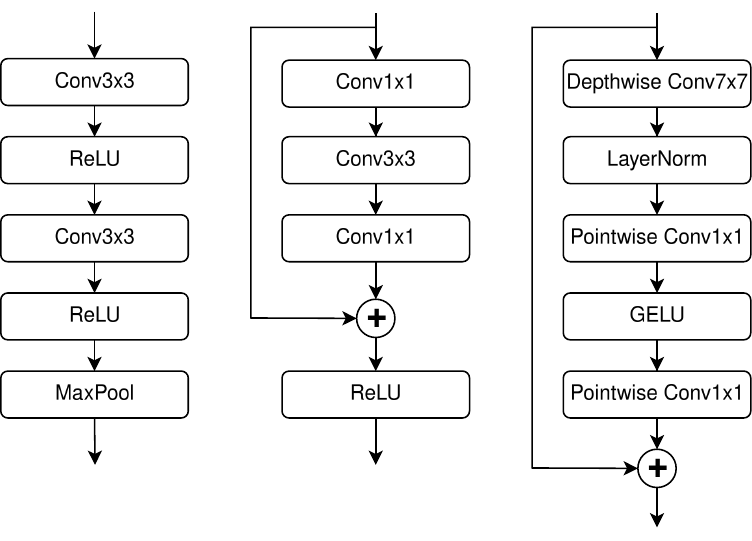} 
  \caption{Fundamental building blocks from VGG16, ResNet50 and ConvNeXt-t architectures (from left to right)}
  \label{fig:3_architectures}
\end{figure}

To evaluate the effect of the model's structural complexity and predictive performance on explainability evaluation metrics, we conduct experiments using three widely studied convolutional neural network architectures of increasing architectural complexity: 
\textbf{VGG16} represents a classical deep network with a uniform structure and no residual connections. \textbf{ResNet50} introduces residual blocks, enabling deeper and more efficient learning by reducing the degradation problem. \textbf{ConvNeXt-t}, a modern architecture inspired by vision transformers, incorporates design choices such as large kernel convolutions and layer normalization, offering a higher degree of model sophistication. A visual comparison of the fundamental building blocks of these architectures is shown in Fig. \ref{fig:3_architectures}, highlighting the progressive increase of architectural complexity.

\noindent For the scope of this study, we focus exclusively on CNN-based architectures, as many computer vision XAI methods were designed for them, whereas transformers would often require specialized XAI approaches.

\newpage \subsection{XAI Methods}
To investigate how a model's structural complexity and accuracy influence different aspects of explainability evaluation, we select two fundamentally distinct XAI methods: Grad-CAM, a feature-attribution-based method, and CRAFT, a concept-based method. 

\noindent \textbf{Gradient-weighted Class Activation Mapping (Grad-CAM)} is a local feature-attribution XAI method designed for CNNs. Grad-CAM assigns “importance” to features based on their contribution to the model’s decision. This “importance” signifies which parts of the input are the most relevant in attaining a selected output. This can be visualized by saliency maps produced by Grad-CAM. \cite{selvaraju2017gradcam}
\\ \noindent \textbf{Concept Recursive Activation FacTorization (CRAFT)} is a concept-based XAI method (CXAI) with a Grad-CAM backbone. CRAFT uses a Non-Negative Matrix Factorization (NMF) to extract concepts from neural network activations and recursively refines them into sub-concepts. It employs so-called Sobol indices to assess each concept's contribution to the model's prediction. The method produces localized visual explanations, concept importance attribution scores, and a hierarchical view of how concepts build up through layers. \cite{fel2023craft}

\subsection{XAI Evaluation Metrics}
To systematically assess the quality of explanations provided by the two aforementioned XAI methods, we adopt multiple metrics which can be assigned to our four chosen key evaluation categories: \textit{Faithfulness}, \textit{Correctness}, \textit{Sensitivity}, and \textit{Simplicity}, each capturing distinct aspects of explanation quality. 

\noindent \textbf{Faithfulness} measures how accurately an explanation reflects a model's decision-making process. \textit{Faithful} explanations highlight features or concepts that, when altered, significantly impact outputs. For Grad-CAM, we use \textit{Faithfulness Correlation} \cite{bhatt2021evaluating}, which measures the correlation between importance scores and prediction changes upon feature removal. For CRAFT, we additionally employ \textit{MoRF AoC} (\textit{Most Relevant First Area over the Curve}) \cite{samek2017evaluating}, which evaluates prediction changes when removing important concepts, and \textit{ROAR AoC} (\textit{Remove and Retrain Area over the Curve}) \cite{hooker2019benchmark}, which assesses model performance after retraining without important concepts. 
\\ \noindent \textbf{Correctness} determines whether an explanation genuinely depends on the model’s learned parameters or could be produced regardless of what the model has learned. \textit{Randomization Tests} compare explanation outputs before and after randomizing model weights. If explanations remain consistent despite randomization, they likely do not reflect the model's internal reasoning. For Grad-CAM, we apply the \textit{Model Parameter Randomization Test (MPRT)} \cite{adebayo2018sanity}, which measures the similarity between the original explanation and that of the perturbed model. Additionally, for both Grad-CAM and CRAFT, we use the \textit{EfficientMPRT (eMPRT)} and \textit{SmoothMPRT (sMPRT)} metrics \cite{hedstrom2024sanity} to further evaluate explanation consistency under weight randomization. 
\\ \noindent \textbf{Sensitivity} evaluates how \textit{robust} explanations are with respect to small, non-adversarial input perturbations. A \textit{robust} explanation method should produce similar results under slight input changes, indicating low volatility. High \textit{Sensitivity} suggests that minor input variations lead to drastically different explanations, undermining reliability. For both Grad-CAM and CRAFT, we use \textit{AvgSensitivity} and \textit{MaxSensitivity} \cite{yeh2019advances}, which measure average and maximum changes in explanations under perturbations, and additionally \textit{SensitivityN} \cite{ancona2018towards} for Grad-CAM, which checks if explanations behave sensitive when modifying several (N) features together.
\\ \noindent \textbf{Simplicity} measures how concise and focused an explanation is, using the fewest meaningful concepts or features to convey the model’s reasoning while avoiding unnecessary complexity or redundancy. For Grad-CAM, we use \textit{Sparseness} \cite{chalasani2020concise}, which quantifies the concentration of importance scores, \textit{Complexity} \cite{bhatt2021evaluating}, which measures the number of features needed for an explanation, and \textit{Effective Complexity} \cite{nguyen2020quantitative}, which evaluates the minimal set of features sufficient for accurate explanation. For CRAFT, we adapt \textit{Complexity} to assess the number of concepts used in explanations.

\section{Experimental Setup}
This section outlines the experimental pipeline, which involves model training, applying XAI methods, and computing evaluation metrics.

\subsection{Model Training}
We trained VGG16, ResNet50 and ConvNeXt-t on each of the 10 subsets from scratch for 200 epochs on a high-performance computing cluster with NVIDIA H100 GPUs with 94GB memory and 768 GB RAM, using PyTorch 2.0 and Python 3.10. All models used a batch size of 32 and cross-entropy loss. Data augmentation included resized cropping to 512x512 pixels, horizontal flipping, and colour jittering. Top-1 accuracy and F1 scores were computed after each epoch of training on the test set.

\subsection{Data Subset and Model Selection}
We selected three subsets (referred to as set 1, 2, and 3 in this paper) for their wide range of top-1 accuracy across architectures, spanning from very low to up to 80\% after some training. For each architecture and subset, we chose 10 model checkpoints at different training stages, representing increasing accuracy and F1 scores, yielding 90 models (3 architectures × 3 subsets × 10 model checkpoints). XAI metrics were aggregated across test set images per model.

\subsection{Generating XAI Evaluations}
In the next step, we applied Grad-CAM and CRAFT on each of the 90 models. For Grad-CAM, explanations were generated for the last convolutional layer of each CNN. For CRAFT, the models were split after the last convolutional layer, using the convolutional backbone as a feature extractor and the remaining layers as a classification head; explanations were likewise produced for each of the 80 classes per set. We then evaluated the explanation outcome with the XAI evaluation metrics introduced in the above section by using the Quantus framework \cite{hedstrom2023quantus} for Grad-CAM and custom adaptations of the metrics for CRAFT.

\section{Results and Discussion}
As described in the above section, we evaluated the explainability of CRAFT and Grad-CAM across 90 model states for three CNN architectures using four XAI metric categories. This section analyzes these metrics in relation to model performance (accuracy ranging from ~10\% to 80\%) and architectural complexity.

\subsection{Faithfulness}
\begin{figure}[ht]
    \centering
    \begin{subfigure}[t]{0.48\textwidth}
        \centering
        \includegraphics[width=\linewidth]{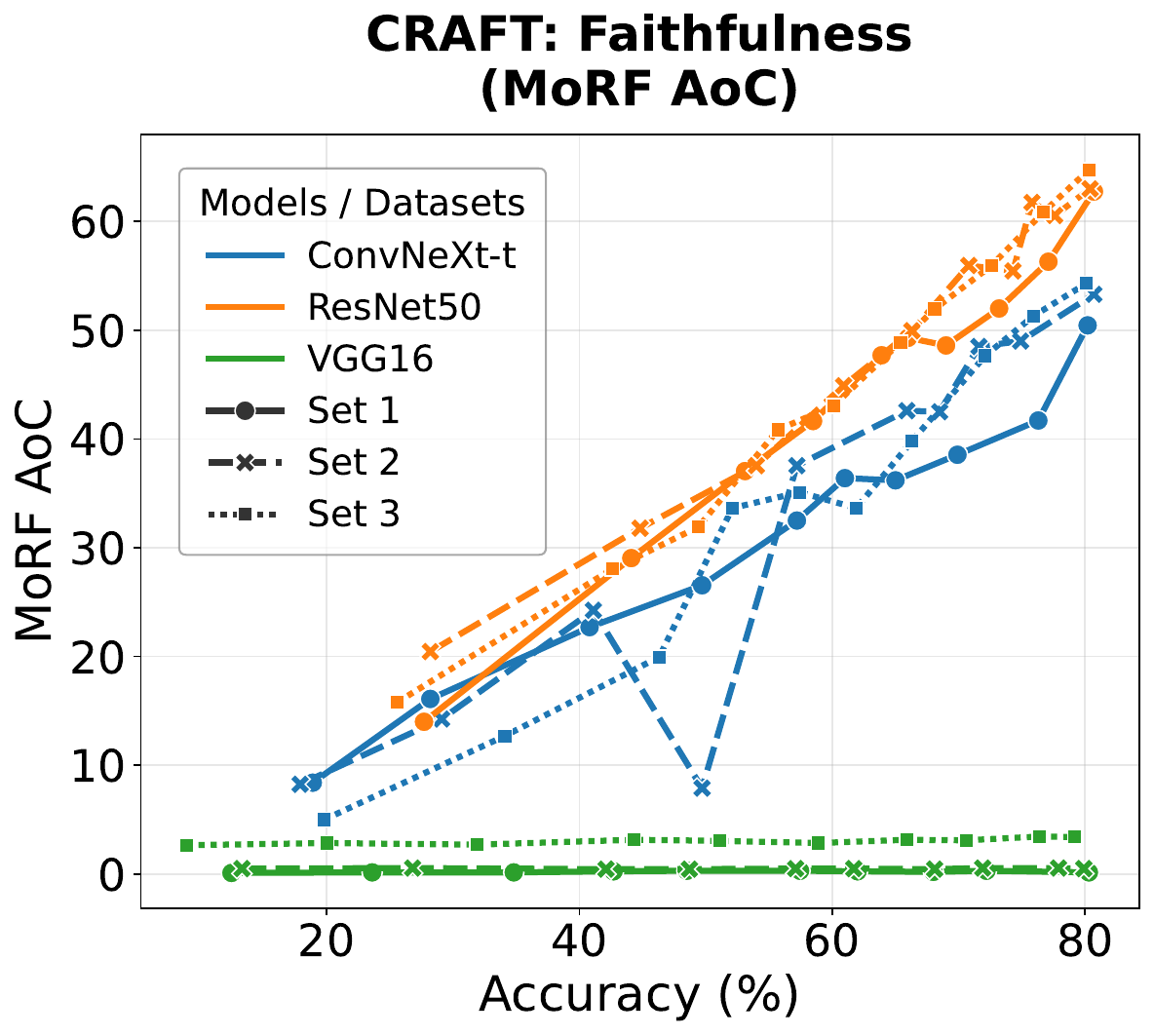}
    \end{subfigure}
    \hfill                                 
    \begin{subfigure}[t]{0.48\textwidth}
        \centering
        \includegraphics[width=\linewidth]{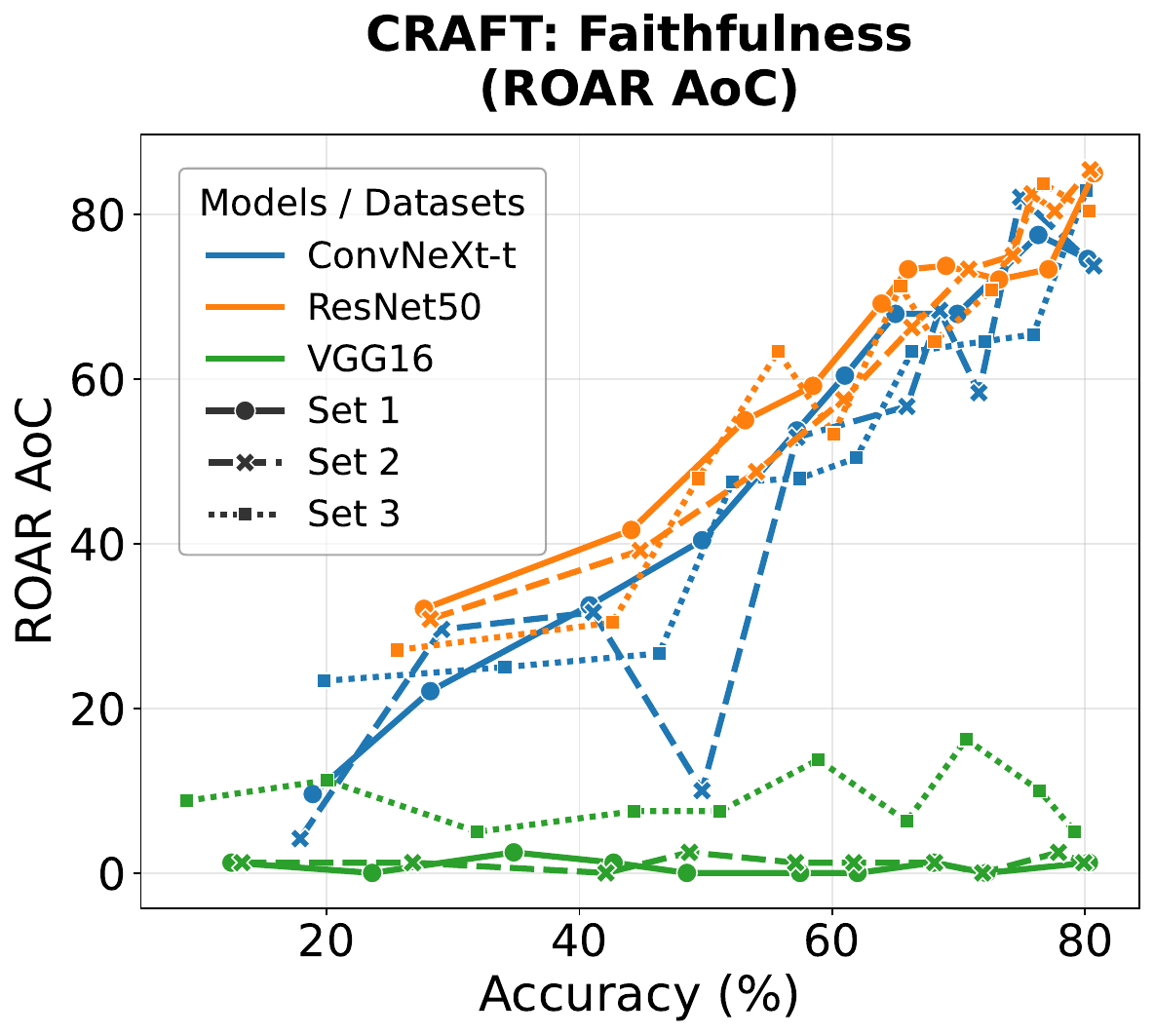}
    \end{subfigure}
    \caption{Faithfulness metrics (\textit{MoRF AoC} and \textit{ROAR AoC}) vs. Accuracy for CRAFT across ConvNeXt-t, ResNet50, and VGG16 on sets 1,2 and 3. Higher values indicate better alignment.} 
    \label{fig:faithfulness}
\end{figure}

\textit{Faithfulness}, captured by \textit{Faithfulness Correlation}, \textit{MoRF AoC} and \textit{ROAR AoC}, measures how well explanations reflect model decisions. Notable tendencies can be seen in Fig. \ref{fig:faithfulness}.
\\ \noindent For both Grad-CAM and CRAFT, \textit{Faithfulness Correlation} showed no consistent trend with accuracy for all three model architectures. However, as shown in Figure \ref{fig:faithfulness}, CRAFT's \textit{MoRF AoC} and \textit{ROAR AoC} increased significantly with accuracy for ConvNeXt-t and ResNet50, but remained relatively low and showed no significant improvement for VGG16.
\begin{itemize}
    \item \textbf{Accuracy vs. Faithfulness}: CRAFT's \textit{MoRF AoC} and \textit{ROAR AoC} increased with accuracy for complex models (ConvNeXt-t, ResNet50), indicating that they learn more nuanced features that explanations can capture with ongoing training. VGG16 showed no meaningful improvement.
    \item \textbf{Architecture vs. Faithfulness}: Complex architectures (ResNet50, Conv\-NeXt-t) produced more \textit{faithful} explanations than VGG16. VGG16's stagnation suggests architectural limitations in producing decision-aligned explanations.
\end{itemize}

\subsection{Correctness}
\begin{figure}[!htb]
    \centering
    \includegraphics[width=1\textwidth]{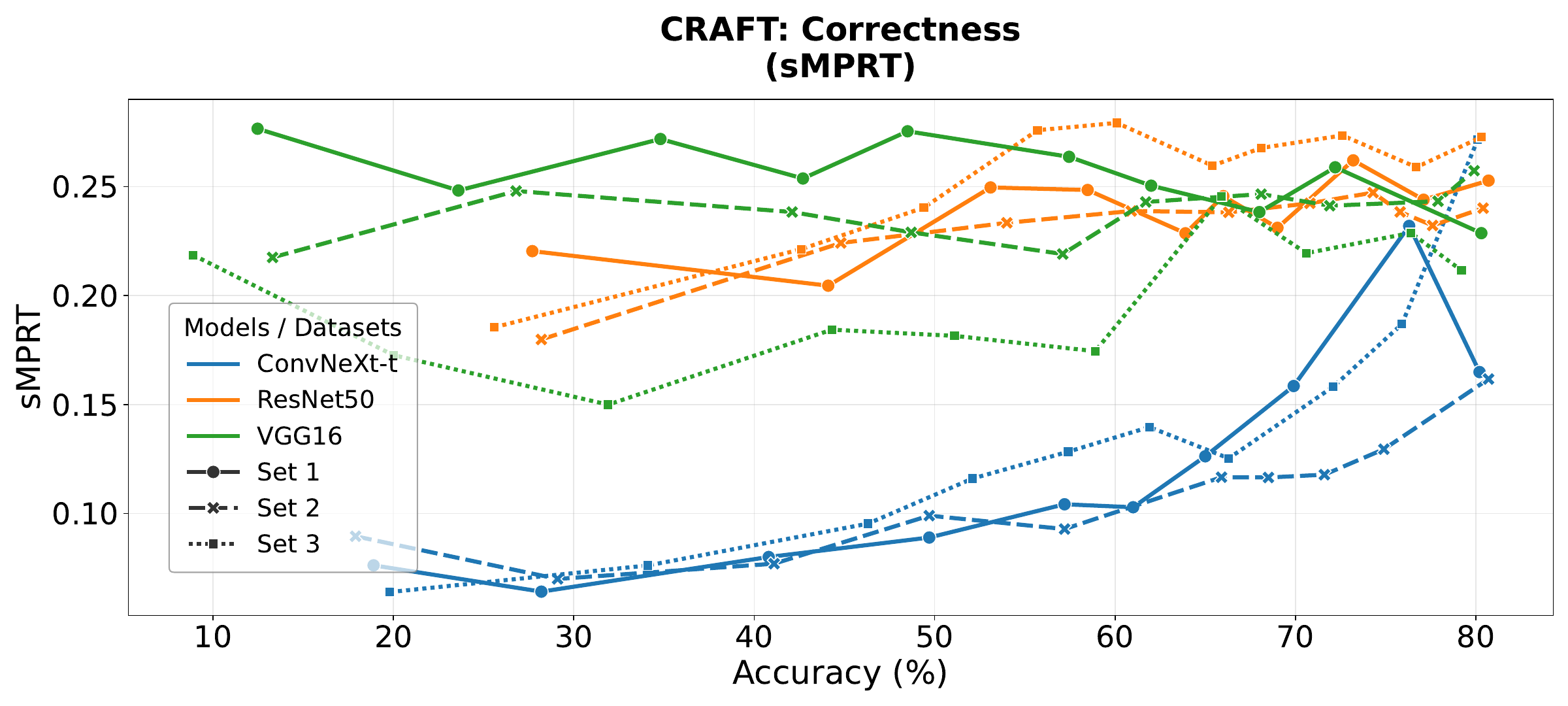}
    \caption{\textit{sMPRT} Score vs. Accuracy for CRAFT across ConvNeXt-t, ResNet50, and VGG16 on sets 1, 2 and 3. Lower scores indicate better \textit{Correctness}.}
    \label{fig:correctness}
\end{figure}

\textit{Correctness} assesses whether explanations depend on learned parameters, with lower \textit{MPRT}, \textit{eMPRT}, and \textit{sMPRT} scores indicating better \textit{Correctness} due to greater changes under weight randomization. Some considerable results are shown in Fig. \ref{fig:correctness}. 
\\ \noindent Grad-CAM’s \textit{MPRT}, \textit{sMPRT} and \textit{eMPRT} values were either zero or consistently very small, with no significant trends observed with respect to accuracy across any architecture.\footnote{This is likely caused by MPRT's denoising problem when applied to gradient-based XAI methods, further explained by \cite{hedstrom2024sanity}. The implication of these findings is simply that these particular XAI metrics are not suited for evaluating Grad-CAM.}

\noindent For CRAFT, \textit{sMPRT} showed a rising trend with accuracy for ConvNeXt-t and ResNet50, indicating worsening \textit{Correctness} at later training stages. No significant trend was observed for VGG16. On average, ConvNeXt-t exhibited the best \textit{Correctness}, especially at early training stages.
\begin{itemize}
    \item \textbf{Accuracy vs. Correctness}: CRAFT's \textit{Correctness} worsened with higher accuracy for ConvNeXt-t and ResNet50, with increasing \textit{sMPRT} scores indicating less sensitivity to model randomization. For VGG16, no clear trend was observed.
    \item \textbf{Architecture vs. Correctness}: VGG16 and ResNet50 showed poorer \textit{Correctness} than ConvNeXt-t, especially during early training stages, as indicated by higher \textit{sMPRT} values. 
\end{itemize}

\subsection{Sensitivity}
\begin{figure}[ht]
    \centering    
    \begin{subfigure}[t]{0.48\textwidth}
        \centering
        \includegraphics[width=\linewidth]{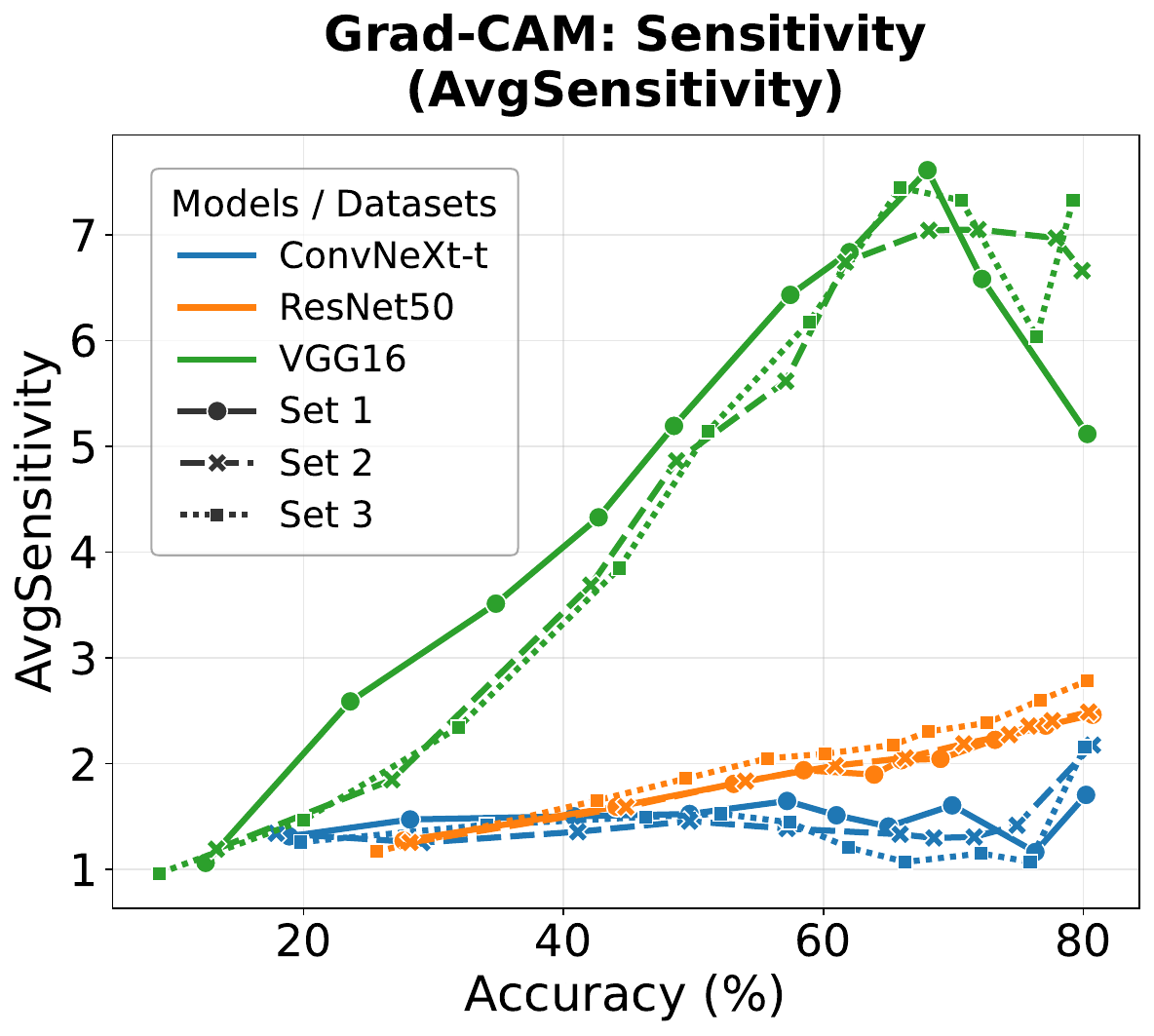}
    \end{subfigure}
    \hfill                                   
    \begin{subfigure}[t]{0.48\textwidth}
        \centering
        \includegraphics[width=\linewidth]{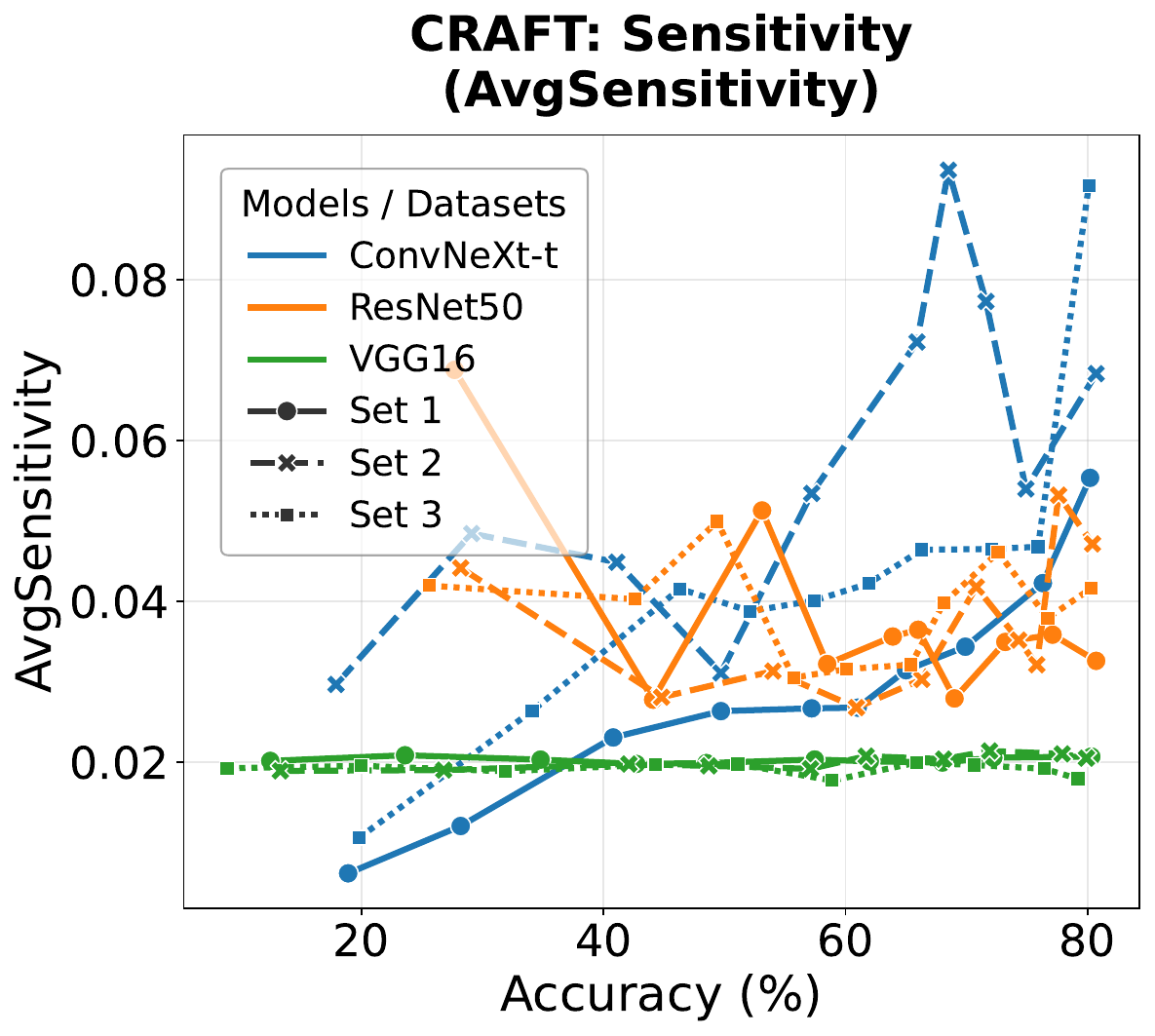}
    \end{subfigure}
    \caption{\textit{AvgSensitivity} vs. Accuracy for Grad-CAM and CRAFT across ConvNeXt-t, ResNet50, and VGG16 on set 1, 2 and 3. Lower \textit{Sensitivity} indicates higher \textit{robustness} against minor input perturbations.}
    \label{fig:sensitivity}
\end{figure}

\textit{Sensitivity} measures explanation stability under small input perturbations using the metrics \textit{AvgSensitivity}, \textit{MaxSensitivity} and \textit{SensitivityN}. Both Grad-CAM's and CRAFT's \textit{MaxSensitivity} are plotted in Fig. \ref{fig:sensitivity}. 
\\ \noindent For Grad-CAM, \textit{AvgSensitivity} strongly increases with model accuracy for VGG\-16 and ResNet50. VGG16 exhibits the highest sensitivity values, followed by ResNet50, while ConvNeXt-t shows the lowest and most stable values, increasing slightly only at a late training stage.\\ 
\noindent CRAFT shows a remarkably different behavior: The aggregated attribution maps show very low sensitivity overall, with VGG16 displaying almost constant and lowest values, Res\-Net50 intermediate values, and ConvNeXt-t the strongest increase with accuracy.

\begin{itemize}
    \item \textbf{Accuracy vs. Sensitivity}: Explanations become more sensitive to small input perturbations as training progresses and accuracy increases. This trend is very pronounced for Grad-CAM across less complex architectures (VGG16 and ResNet50). For CRAFT, only ConvNeXt-t shows an increasing trend of Sensitivity with growing accuracy.
    \item \textbf{Architecture vs. Sensitivity}: ConvNeXt-t produces the most robust (least sensitive) explanations for Grad-CAM, followed by ResNet50, while VGG16 shows by far the highest sensitivity, indicating that its explanations rely on perturbation sensitive features. For CRAFT the aggregated attribution maps demonstrate very low sensitivity compared to Grad-CAM, indicating more robust explanations due to the heavy smoothing through global NMF concept extraction, spatial averaging, and weighted linear combination of multiple concept heatmaps. VGG16 exhibits the most stable (lowest) sensitivity, followed by ResNet50, with ConvNeXt-t showing the largest sensitivity growth and highest values at high accuracy.
\end{itemize}

\subsection{Simplicity}
\begin{figure}[ht]
    \centering
    \begin{subfigure}[t]{0.48\textwidth}
        \centering
        \includegraphics[width=\linewidth]{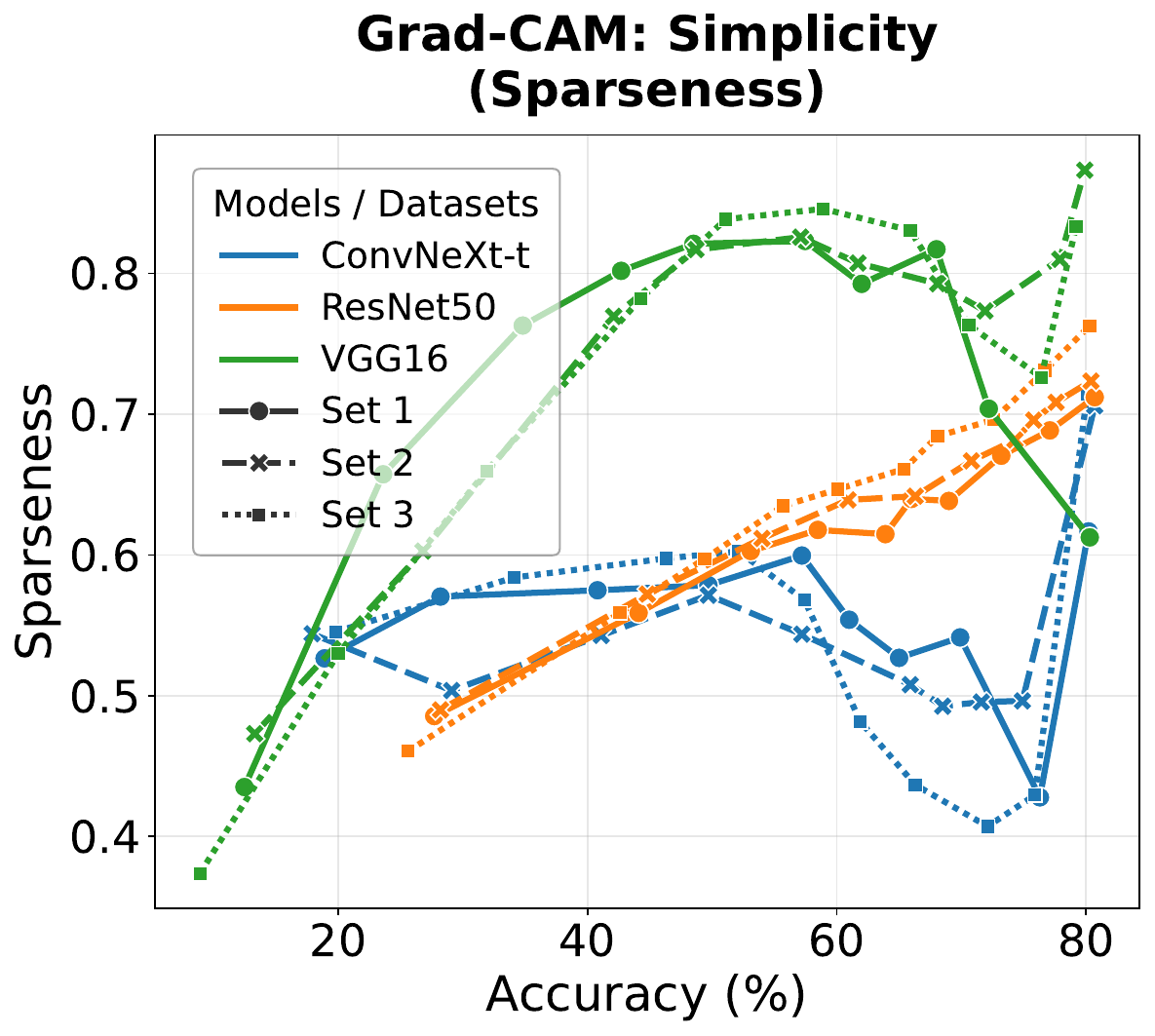}
    \end{subfigure}
    \hfill                                   
    \begin{subfigure}[t]{0.48\textwidth}
        \centering
        \includegraphics[width=\linewidth]{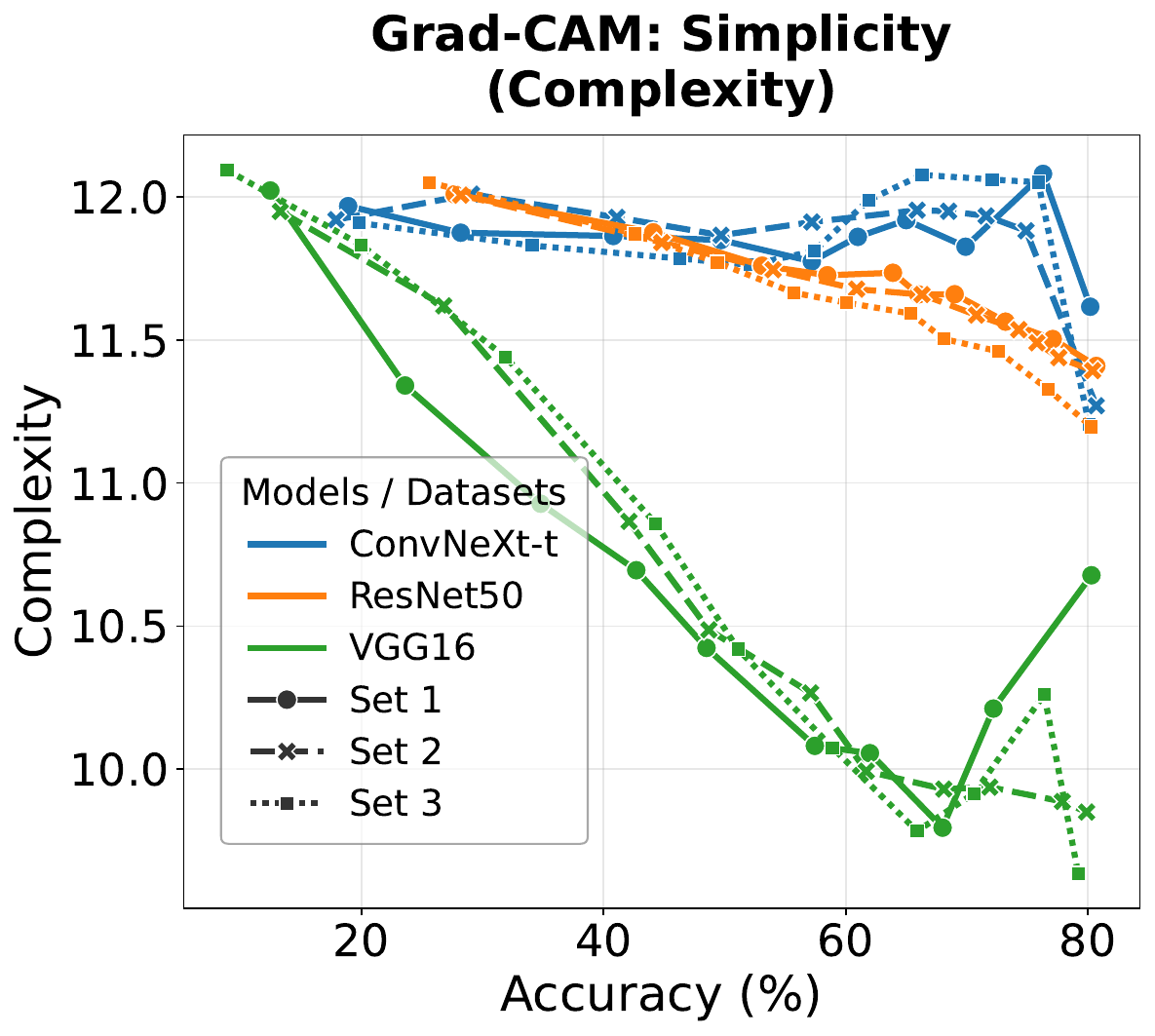}
    \end{subfigure}
    \caption{\textit{Sparseness} and \textit{Complexity} vs. Accuracy for Grad-CAM across ConvNeXt, ResNet50, and VGG16 on sets 1,2 and 3. Higher \textit{Sparseness} and lower \textit{Complexity} indicate focused explanations.}
    \label{fig:simplicity}
\end{figure}

\textit{Simplicity} evaluates explanation focus and conciseness, measured by the metrics \textit{Complexity}, \textit{Effective Complexity} and \textit{Sparseness}. Notable results are plotted in Fig. \ref{fig:simplicity}. 

\noindent
For Grad-CAM, \textit{Sparseness} increases substantially with accuracy for VGG16 and, to a lesser extent, for ResNet50. ConvNeXt shows only weak and inconsistent variation in sparseness, with no clear trend. The \textit{Complexity} metric confirms these patterns: VGG16 exhibits the strongest decrease, ResNet50 shows a moderate decline, while ConvNeXt remains largely stable with only a minor drop at the highest accuracies. \textit{Effective Complexity} shows high variability without consistent trends across architectures and was therefore not emphasized further.
\noindent For CRAFT, the \textit{Complexity} metric remains very high and remarkably stable across all architectures and accuracy levels, with only marginal decreases visible at the highest accuracies.
\begin{itemize}
    \item \textbf{Accuracy vs. Simplicity}: For Grad-CAM, explanations become remarkably sparser and less complex as training progresses and accuracy increases, especially for VGG16 and (to a lesser degree) ResNet50. ConvNeXt-t shows only limited improvement in simplicity. CRAFT explanations remain consistently complex with virtually no change over training.
    \item \textbf{Architecture vs. Simplicity}: VGG16 consistently produces the sparsest and least complex Grad-CAM explanations, particularly at lower accuracies. ResNet50 shows average performance, while ConvNeXt-t explanations remain the least sparse and most complex overall, suggesting that architectural complexity may encourage reliance on distributed or fine-grained features. CRAFT explanations are uniformly complex regardless of architecture.
\end{itemize}

\section{Conclusion and Future Work}
This study explores the interplay between model accuracy, architectural complexity, and the quality of explanations produced by XAI methods, offering insights into their relationships within computer vision applications.

\noindent We assessed four XAI evaluation categories in relation to model accuracy and architectural complexity across VGG16, ResNet50, and ConvNeXt-t. While our evaluations we conducted in this study reveal no consistent dependency across all metrics, some plausible correlations emerged:
\begin{enumerate}
    \item Complex architectures like ResNet50 and ConvNeXt-t generally produced more \textit{faithful} explanations, particularly as accuracy increased, aligning with expectations of better feature alignment in advanced models. These models also exhibited lower \textit{Sensitivity} and higher \textit{Correctness}, suggesting more robust and parameter-dependent explanations.
    \item Conversely, VGG16’s simpler architecture yielded less \textit{Faithfulness} and less \textit{stable} explanations with higher \textit{Sensitivity} and poorer \textit{Correctness}. In terms of \textit{Simplicity}, VGG16's Grad-CAM explanations outperformed the rather complex architectures significantly, especially after a few epochs of training.
\end{enumerate} 
\vspace{10pt}
These findings highlight trade-offs in selecting XAI methods and model architectures, guiding their application based on the prioritized balance of explanatory quality and model accuracy.

\noindent While these results provide a foundation for understanding how model architecture and accuracy influence XAI explanation quality, the study’s scope was constrained by its focus on specific models, datasets, and methods. To build on these insights and address these limitations, further exploration is needed to validate and extend our observations across more diverse contexts and evaluation frameworks. Future research should encompass a broader range of model architectures, as well as diverse datasets beyond ImageNet-1K, to generalize these findings. Additionally, evaluating a wider spectrum of XAI methods, including other feature-attribution-based approaches like LIME and SHAP, and concept-based methods like CRP \cite{achtibat2023attribution}, would enhance our understanding of explanation quality across varied model contexts. Incorporating more XAI evaluation metrics could further validate and extend our observations, enhancing the applicability of XAI in high-risk domains.

\begin{credits}
\subsubsection{\ackname} This study was partially supported by DFG (grant number TE257/37-1), by the Konrad Zuse School, SECAI (BMBF Project Nr. 57616814) and by the DZSF as part of the XRAISE project. The authors gratefully acknowledge the computing time made available to them by ZIH at TU Dresden. 

\subsubsection{\discintname}
The authors have no competing interests to declare that are
relevant to the content of this article.
\end{credits}
%
%

\bibliographystyle{splncs04}
\bibliography{literature}

\end{document}